\documentclass[journal]{IEEEtran}
\usepackage[T1]{fontenc}
\usepackage{booktabs}
\usepackage{multirow,graphicx}
\usepackage{amsmath}
\usepackage{color}
\usepackage{subfigure}
\usepackage{amssymb}
\usepackage{array}%
\usepackage{makecell}
\usepackage{threeparttable}
\usepackage{amsmath,graphicx,multirow,algorithm,algorithmic,bm}
\usepackage[switch,pagewise,columnwise]{lineno}
\usepackage[colorlinks,linkcolor=black]{hyperref}
\makeatletter
\let\NAT@parse\undefined
\makeatother
\usepackage{hyperref}

\ifCLASSINFOpdf
\else
\fi
\usepackage{amsmath}
\begin{document}
	%
	\title{GLSD: A Global Large-Scale Ship Database with Baseline Evaluations}
	%
	%
	%
	
	\author{Zhenfeng Shao, Jiaming Wang, Lianbing Deng, Xiao Huang, Tao Lu, Fang Luo, \\
		 Ruiqian Zhang, Xianwei Lv, Chaoya Dang, Qing Ding, and Zhiqiang Wang
	\thanks{
	Z. Shao, J. Wang, X. Lv, Q. Ding, C. Dang, and Z. Wang are with the State Key Laboratory for Information Engineering in Surveying, Mapping and Remote Sensing, Wuhan University, Wuhan 430079, China (e-mail: {shaozhenfeng, wjmecho, xianweilv, dingqing}@whu.edu.cn, chaoyadang99@163.com, wangzqwhu@foxmail.com). Z. Shao is with State Key Laboratory of Severe Weather, Chinese Academy of Meteorological Sciences, Beijing, 100081, China.

	L. Deng is with the Post-Doctoral Research Center of Zhuhai Da Hengqin Science and Technology Development Co., Ltd, Guangdong Hengqin New Area, 519031, China (e-mail: denglb@csu.edu.cn).
	
	X. Huang is with the Department of Geosciences, University of Arkansas, Fayetteville, AR, 72701, USA (e-mail: xh010@uark.edu).
	
	T. Lu is with the School of Computer Science and Engineering, Wuhan Institute of Technology, Wuhan 430205, China (e-mail: lutxyl@gmail.com).
	
	F. Luo is with the School of Computer and Artificial Intelligence, Wuhan University of Technology, Wuhan 430063, China; (e-mail: luof@whut.edu.cn).
	
	R. Zhang is with the Chinese Academy of Surveying and Mapping, Beijing 100036, China (e-mail: zhangruiqian@whu.edu.cn).

	This work was supported in part by the National Key R\&D Program of China under Grant 2018YFB0505401; in part by the National Natural Science Foundation of China under Grants 41890820, 41771452, 41771454, and 62072350; in part by the Key R\&D Program of Yunnan Province in China under Grant 2018IB023; in part by the Open Grants of the State Key Laboratory of Severe Weather under Grants 2021LASW-A17 (Corresponding author: Jiaming Wang and Tao Lu).
	} 

	}

	\markboth{Journal of \LaTeX\ Class Files,~Vol.~14, No.~8, August~2015}%
	{Shell \MakeLowercase{\textit{et al.}}: Bare Demo of IEEEtran.cls for IEEE Communications Society Journals}

	\maketitle
	
	\begin{abstract}
	In this paper, we introduce a challenging global large-scale ship database (called GLSD), designed specifically for ship detection tasks. The designed GLSD database includes a total of 212,357 annotated instances from 152,576 images. Based on the collected images, we propose 13 ship categories that widely exist in international routes. These categories include \textsl{Sailing boat, Fishing boat, Passenger ship, Warship, General cargo ship, Container ship, Bulk cargo carrier, Barge, Ore carrier, Speed boat, Canoe, Oil carrier, and Tug}. The motivations of developing GLSD include the following: 1) providing a refine and extensive ship detection database that benefits the object detection community, 2) establishing a database with exhaustive labels (bounding boxes and ship class categories) in a uniform classification scheme, and 3) providing a large-scale ship database with geographic information (covering more than 3000 ports and 33 routes) that benefits multi-modal analysis. In addition, we discuss the evaluation protocols corresponding to image characteristics in GLSD and analyze the performance of selected state-of-the-art object detection algorithms on GSLD, aiming to establish baselines for future studies. More information regarding the designed GLSD can be found at \url{https://github.com/jiaming-wang/GLSD}.
	\end{abstract}
	

	\IEEEpeerreviewmaketitle
	
	\section{Introduction}
 
Object detection has been an important computer vision task for over 20 years \cite{zou2019object}. In recent years, with the growing demand for public security, the detection of ships has become an important task in both military and civilian fields \cite{shao2018seaships}, including sea controlling, illegal smuggling monitoring, and automatic driving. The rapid development of artificial intelligence also pushes autonomous ship detection to the spotlight. Ship detection is of great importance, as sea routes are the lifeblood of the global economy \footnote{\url{https://www.ics-shipping.org/shipping-facts/shipping-and-world-trade}}, given the fact that the international shipping industry is responsible for the carriage of around 90\% of world trade . Nevertheless, manual inspection for identifying abnormal behaviors is a time-consuming and laborious process. With the development of the ship automatic navigation system, the demand for a gigantic amount of data for data-driven models is rising. In addition, despite that automatic object detection methods have achieved great performance, it is still far from maturity, as challenges still remain when those algorithms are being applied in real-world ship detection scenarios.

Inspired by the immense success of machine learning approaches in computer vision tasks \cite{lu2020global, ma2020scscn, ma2019fusiongan,wang2021spatial}, deep learning-based methods have been the mainstream in addressing object detection problems \cite{zou2019object}. However, the performance of deep-learning-based algorithms, given their big-data-driven nature \cite{xia2018dota}, largely depends on the number of high-quality training samples. The first large-scale dataset, i.e., ImageNet \cite{deng2009imagenet}, has been widely adopted in object detection studies and even other vision tasks \cite{ledig2017photo,wang2021dual}. Following ImageNet, Lin \emph{et al.} \cite{lin2014microsoft} presented the Microsoft common objects in context (MS COCO) dataset with instance-level segmentation masks. In real-world scenarios, ships with different categories play considerably different roles during sea transportation. In these publicly available datasets, however, ships are commonly generalized as ``ship'' or ``boat'' (for example, in VOC2007 \cite{everingham2010pascal}, CIFAR-10 \cite{krizhevsky2009learning}, Caltech-256 \cite{griffin2007caltech}, and COCO \cite{lin2014microsoft}).  Although, ImageNet \cite{deng2009imagenet} includes six types of ships, i.e., ``fireboat'', ``lifeboat'', ``speedboat'', ``submarine'', ``pirate'', and ``container ship'', most of them are seagoing vessels that are rarely seen in certain situations. Thus, we argue that object models learned from the aforementioned coarse-grained datasets are not suitable for ship identification and the corresponding applications in real-world scenarios. Developing a new large-scale ship database is of great necessity.

	\begin{table*}[htbp]
	 	\centering
	 	\renewcommand\arraystretch{1.2}
	 	\caption{Comparison of ships among GLSD and object detection datasets. Improving from the first ship dataset SeaShips, we add the following categories: ``Sailing boat'', ``Warship'', ``Barge'', ``Speed boat'', ``Canoe'', ``Oil carrier'', and ``Tug''. GLSD have a total of 152,576 images.}
	 	\begin{tabular}{c||ccccc}
	 		\hline
	 		Dataset & main categories & instances & images & image size & Boxes/img\\
	 		\hline \hline
	 		VOC2007 \cite{everingham2010pascal} & 1 & 791   & 363   & random &2.18\\
	 		CIFAR-10 \cite{krizhevsky2009learning} & 1 & 6,000  & 6,000  & 32 $\times$ 32 &1.00\\
	 		Caltech-256 \cite{griffin2007caltech} & 4     & 418   & 418   & random &1.00\\
	 		ImageNet \cite{deng2009imagenet} & 1     & 613 & 525 & random &1.17\\
	 		COCO \cite{lin2014microsoft} & 1     & 10,759 & 3,025 & random &3.56\\
	 		SeaShips \cite{shao2018seaships} & 6     & 40,077 & 31,455 & 1,920 $\times$ 1,080 &1.37\\
	 		\hline \hline
	 		GLSD  & 13    & 212,357   & 152,576  & random & 1.39\\
	 		\hline
	 	\end{tabular}%
	 	\label{tabl}%
	 \end{table*}%
 
	 Recently, efforts have been made to construct ship datasets. For instance, Shao \emph{et al.} \cite{shao2018seaships} developed  a ship dataset, i.e., SeaShips, that consists of 31,455 high-quality images (1,920 $\times$ 1,080 pixels) and covers six common ship types (i.e., ``passenger ship'', ``fishing boat'', ``container ship'', ``general cargo ship'', ``bull cargo carrier'', and ``ore carrier''). Despite the fact that the SeaShip dataset considers the following factors: various scales, viewpoints, backgrounds, illumination, and diverse occlusion conditions, these ship images were derived from cameras that only cover the Zhuhai Hengqin New Area, China, with considerably simple scenes and an unbalance category distribution (e.g., a large number of fishing boats but limited numbers in other types). Another notable effort is by Zheng \emph{et al.} \cite{zheng2020mcships} who presented a new multi-category ship dataset, namely McShips, that aimed at the fine-grained categorization of both civilian ships and warships. In McShips, warships are divided into six sub-categories. However, The McShips dataset is relatively small in size, and the ratio of the number of civilian ships to the number of warships is roughly 1:1, which is not in accordance with the real-world scenario where there are far more civilian ships than warships. As McShips is not yet publicly available, we do not intend to further discuss it in this study.

	In this study, we present a novel ship dataset, called the Global Large-Scale Ship Database (GLSD), that consists of 152,576  images and covers 13 ship types. Considering that the routes of ships are well established, we collect internet images and monitoring data according to the routes with port and country information. The developed GLSD covers more than 3,000 ports around the world (more details in Section \ref{s31}). Improving from SeaShips, we add the following categories: ``Sailing boat'', ``Warship'', ``Barge'', ``Speed boat'' ``Canoe'', ``Oil carrier'', and ``Tug''. Labels and bounding boxes of GLSD are manually constructed in an accurate manner using an image annotation tool \cite{russell2008labelme}.  We name the route-based version of GLSD as ``GLSD\_port'' (GLSD with geographic information, more details about the ``GLSD\_port''  in \url{https://github.com/jiaming-wang/GLSD/blob/master/Ports\_list.md}). We believe that GLSD\_port provides training models with multi-modal information that potentially benefits certain ship detection applications.
	 
	 Detailed comparisons of GLSD with existing databases are shown in Table \ref{tabl}. The early database, like CIFAR-10 \cite{krizhevsky2009learning}, has a very low image resolution (32 $\times$ 32 pixels), which is not suitable for object detection tasks. The maximum number of boxes per image in the PASCAL VOC2007 \cite{everingham2010pascal}, Caltech-256 \cite{griffin2007caltech}, and ImageNet \cite{deng2009imagenet}, is rather small. Although COCO \cite{lin2014microsoft} includes a great quantity of ``boat'' images, the ``boat'' category has not been sub-categorized. Compared with SeaShips, the proposed GLSD contains more categories (13 vs. 6). In terms of image quantity, the number of images in our GLSD is three times compared to the number of images in SeaShips. In addition, GLSD owns a larger number of boxes per image than the one in SeaShips (1.39 vs. 1.37). Table \ref{tabl} indicates that GLSD is a more challenging ship database that potentially benefits the training of robust ship detection models.
	 

	The main contributions of this work are summarized as follows: 
	\begin{enumerate}
		\item To our best knowledge, the developed GLSD is a very challenging global ship dataset with the largest number of annotates, reaching above 150,000 images, potentially facilitating the development and evaluation of existing object detection algorithms. 
		\item GLSD is built on global routes, providing multi-modal information (port and country of image acquisition) that better serves certain ship detection tasks. We plan to maintain GLSD in a regular manner when new images are available. 
		\item We evaluate state-of-the-art object detection algorithms on the proposed GLSD, setting up a baseline for future algorithm development.
	\end{enumerate}
	
	This paper is organized as follows. Section \ref{s2} reviews the related work. Sections \ref{s3} and \ref{s4} illustrate the collection and design of the GLSD database. Section \ref{s6} details the experiments and analysis. Section \ref{s7} concludes this study.

	\section{Related Work} \label{s2}
	In this section, we outline the development of object detection datasets and methods, providing references for future studies.
	
	\subsection{Object Detection Datasets}
	
	In the early days, some small-scale well-labeled datasets (i.e., Caltech10/256 \cite{fei2006one,griffin2007caltech}, MSRC \cite{shotton2006textonboost}, PASCAL \cite{everingham2015pascal}, and CIFAR-10 \cite{krizhevsky2009learning}) were widely used in computer vision tasks as benchmarks. These datasets offer a limited number of categories with low-resolution images (such as, 32 $\times 32$ and $300 \times 200$ pixels) \cite{deng2009imagenet}.

	It is widely acknowledged that the development of deep learning is inseparable from the support of big data. In general,  high-quality training data can lead to the better performance of the deep-learning algorithms. For the first time, Deng \emph{et al.} \cite{deng2009imagenet} built a dataset with worldwide targets following a tree structure organization, pushing object classification and detection fields towards more complex problems. The dataset proposed by Deng \emph{et al.} \cite{deng2009imagenet} now contains 14 million images that cover 22 categories. Later on, pre-training backbone networks \cite{simonyan2014very, he2016deep} based on the ImageNet images gradually became the benchmark in computer vision tasks. From early datasets, like COCO \cite{lin2014microsoft}, to the recent benchmarks, like Objects365 \cite{shao2019objects365}, large-scale datasets have always been preferred choices by deep learning algorithm developers, as they play an essential role in evaluating the performance of object classification and detection tasks.

	Besides the above general object detection datasets, many datasets have been developed for specific scenarios, e.g., masked face recognition for novel coronavirus disease 2019 (COVID-19)  pneumonia (RMFD \cite{wang2020masked}), music information retrieval (GZTAN \cite{tzanetakis2002musical} and MSD \cite{bertin2011million}), automated detection and classification of fish (labeled fishes \cite{cutter2015automated}), autonomous driving (JAAD \cite{rasouli2017ICCVW} and LISA \cite{sivaraman2010general}), and ship action recognition database \cite{wang2021spatial}. These domain-specific datasets have greatly facilitated the development of  the corresponding tasks and applications. In a recent effort, Shao \emph{et al.} \cite{shao2018seaships} constructed the first large-scale dataset for ship detection, i.e., SeaShips. Due to the fixed viewpoint in the deployed video monitoring system in the Zhuhai Hengqin New Area, however, the background information in SeaShips lacks diversity. Another notable effort is by Zheng \emph{et al.} \cite{zheng2020mcships}, who presented a new multi-category ship dataset, i.e., McShips. However, ship targets in McShips are with an unreasonable ratio among different ship categories.

	\subsection{Object Detection Algorithms}
	
	An object detection model generally consists of two main components, a backbone pre-trained on a large image dataset (e.g., ImageNet \cite{deng2009imagenet}) for feature extraction and a head used to predict the label. Common backbone networks include VGG \cite{simonyan2014very}, ResNet \cite{he2016deep}, DenseNet \cite{huang2017densely}, and ResNetXt \cite{xie2017aggregated}. Existing head components can be divided into two categories, i.e., the traditional methods and deep-learning methods \cite{zou2019object}. In the early stages, most object detection methods adopted hand-crafted image features to achieve real-time object detection \cite{viola2001rapid, viola2001robust}. Histogram of oriented gradients (HOG) detector \cite{felzenszwalb2008discriminatively} played a very important role in this task. Felzenszwalb \emph{et al.}  \cite{felzenszwalb2008discriminatively} proposed a deformable part-based model (DPM) that can be viewed as an extension of the HOG detector. DPM \cite{felzenszwalb2008discriminatively} gradually became the main theme of pedestrian detection as a pretreatment \cite{zheng2015scalable}.

	According to the network structure, deep-learning-based object detection methods can be grouped into two genres: two-stage and one-stage detection \cite{zou2019object}, where the two-stage detectors are the dominant paradigm of the object detection tasks. Girshick \emph{et al.} \cite{girshick2015region} proposed the regions with convolutional neural networks (CNN) feature maps for object detection, establishing a brand new venue for the development of two-stage detection algorithms. To reduce computational complexity, SPP-Net \cite{he2014spatial} largely reduced the computing cost through the spatial pyramid pooling layer. Inspired by SPP-Net, Girshick \emph{et al.} \cite{girshick2015fast} further utilized a more efficient region of interest (ROI) pooling to reduce unnecessary computational overhand. In 2015, Ren \emph{et al.} \cite{ren2015faster} first proposed a framework that introduces a region proposal network to obtain bounding boxes with low complexity. Lin \emph{et al.} \cite{lin2017feature} proposed the feature pyramid network (FPN), which fused multi-scale features to enhance semantic information expression.

	Different from the above deep learning-based algorithm, YOLO \cite{redmon2016you} transforms the detection and classification into an end-to-end regression model, sacrificing the localization accuracy for a boost of detection speed. In the following development of the YOLO series \cite{redmon2017yolo9000, redmon2018yolov3, bochkovskiy2020yolov4}, its subsequent versions inherit its advantages, while trying to gradually improve the detection accuracy. Liu \emph{et al.} \cite{liu2016ssd} proposed a multi-reference and multi-resolution framework that can significantly enhance detection accuracy. Further, Lin \emph{et al.} \cite{lin2017focal} introduced the focal loss the prevents the accuracy drop resulting from the imbalance foreground-background classes in one-stage detection methods.

		\begin{table*}[htbp]
		\centering
		\renewcommand\arraystretch{1.1}
		\caption{The definition of main categories in GLSD.}
		\begin{tabular}{c|c}
			\hline
			Categories & Definition \\
			\hline \hline
			Ore carrier & Ships with small stowage factors \\
			Bulk cargo carrier & Ships with large stowage factors \\
			General cargo ship & Ships that transport machinery, equipment, building materials, daily necessities, etc. \\
			Container ship & Ships that specialize in the transport of containerized goods \\
			Fishing boat & Ships that catch and harvest aquatic animals and plants \\
			Passenger ship & Vessels used to transport passengers or pedestrians \\
			Sailing boat & Boats propelled partly or entirely by sails \\
			Barge & Ships for cargo transportation between inland ports \\
			Warship & Naval ships that are built and primarily intended for naval warfare \\
			Oil carrier & Ships designed for the bulk transport of oil or its products\\
			Tug & Tug maneuvers other vessels by pushing or pulling them either by direct contact or by means of a tow line\\
			Canoe & Lightweight narrow vessels typically pointed at both ends and open on top \\
			Speed boat & Small boats with a powerful engine that travels very fast  \\
			\hline
		\end{tabular}%
		\label{tab2}%
	\end{table*}%

	\section{Image Collection} \label{s3}
   
   
   In this section, we present the details on the collection, main categories, and characteristics of the GLSD.
   
	\subsection{Port-based image collection} \label{s31}
	Referring to the United Nations Code for Trade and Transport Locations (UN/LOCODE)\footnote{\url{http://www.unece.org/cefact/locode/welcome.html}}, global ports are divided into 33 routes, i.e., ``east of South America'', ``Pacific island'', ``West Mediterranean'', ``Middle East'', ``Caribbean'', ``West Africa'', ``Australia'', ``India-Pakistan'', ``European basic port'', ``European inland port'', ``East Mediterranean'', ``Black Sea'', ``Southeast Asia'', ``Canada'', ``west of South America'', ``China'', ``Taiwan-China'', ``East Africa'', ``North Africa'', ``Red Sea'', ``partial port of Japan'', ``Adriatic Sea'', ``Kansai'', ``Kanto'', ``Korea'', ``Mexico'', ``South Africa, ``New Zealand'', ``west of American'', ``Russia Far East'', ``American inland port'', ``east of American'', and ``Central Asia''.

	To ensure diverse image sources, we try to collect images from as many ports as possible (the ports involved in the dataset can be found on our website\footnote{\url{https://github.com/jiaming-wang/GLSD/blob/master/Ports\_list.md}}). A certain number of images in GLSD are captured from a deployed video monitoring system in the Zhuhai Hengqin New Area, China, and the rest are collected via search engines with multiple resolutions. As images in certain routes are unavailable, GLSD mainly covers ship images captured in China, America, and Europe.

	\begin{figure}[h]
		\centering
		\includegraphics[height=6cm]{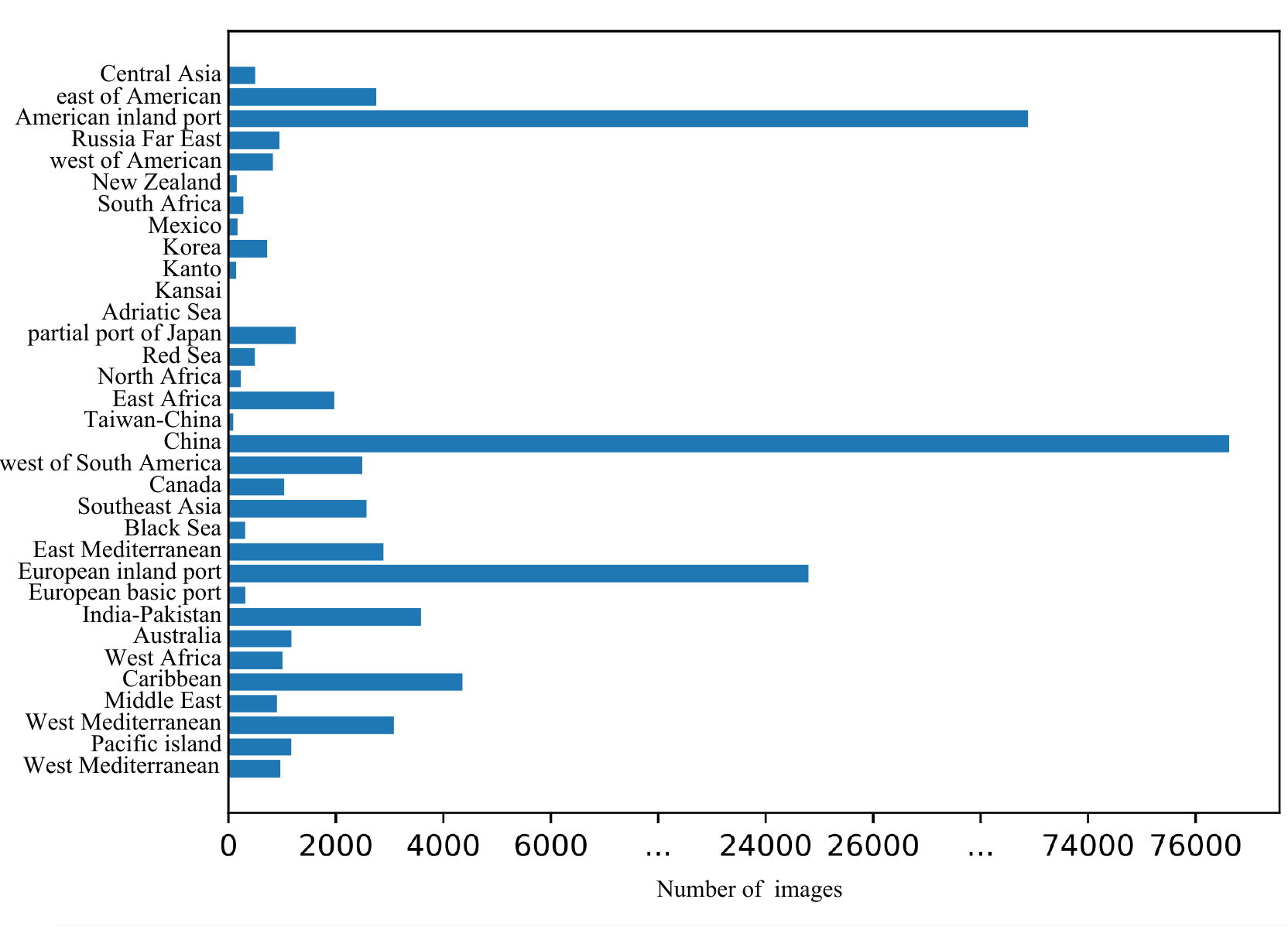}
		\caption{The distribution of ports covered by GLSD.}
		\label{routes}
	\end{figure}

	\subsection{Iconic and non-iconic Image Collection}
\begin{figure}[h]
		\centering
		\includegraphics[height=8.5cm]{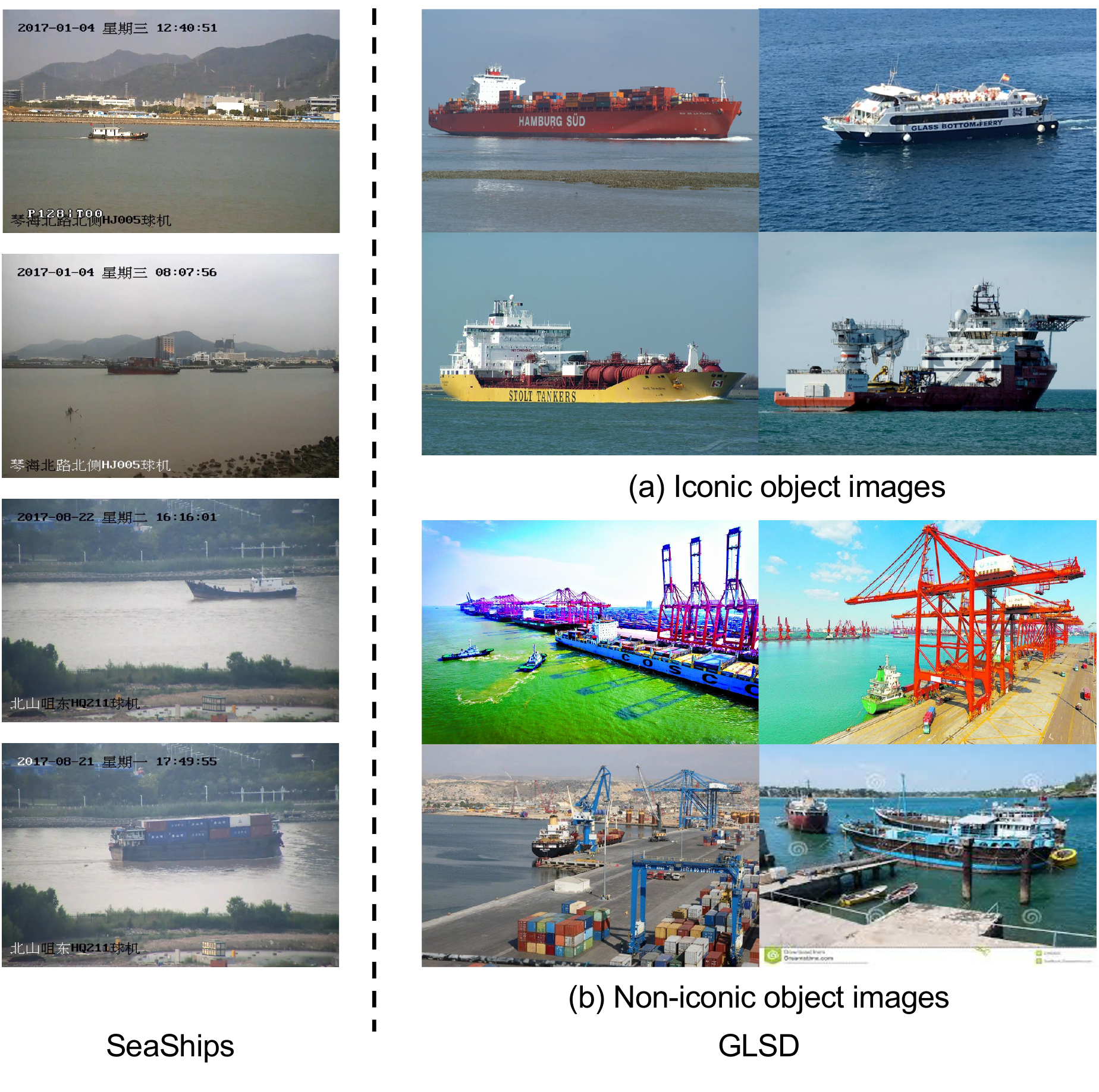}
		\caption{Selected examples of (a) iconic object images and (b) non-iconic object images from the GLSD.}
		\label{iconic}
	\end{figure}

	Images in the GLSD can be roughly divided into two categories: iconic images \cite{berg2009finding} and non-iconic images \cite{lin2014microsoft}. Iconic images, often with a clear depiction of categories, provide high-quality object instances, which clearly depict objects' categories (see examples in Fig. \ref{iconic}(a)). Iconic images are widely used in object classification and retrieval tasks, and they can be directly retrieved via image search engines. Most images in SeaShips are iconic images. Non-iconic images that provide contextual information and non-canonical viewpoints also play an important role in object detection tasks (see examples in Fig. \ref{iconic}(b)). In the proposed GLSD, we keep both iconic and non-iconic images, aiming to provide diverse image categories that benefit object detection model training. Compared with SeaShips that contain mostly iconic images, GLSD is considerably more challenging and closer to real-world scenes.
	
	\subsection{Main Categories}
	From collected images, we construct 13 categories that widely exist in international routes. These categories include \textsl{``Sailing boat'', ``Fishing boat'', ``Passenger ship'', ``Warship'', ``General cargo ship'', ``Container ship'', ``Bulk cargo carrier'', ``Barge'', ``Ore carrier, ``Speed boat'', ``Canoe'', ``Oil carrier'', and ``Tug''}. We refer to Wikipedia\footnote{\url{https://www.wikipedia.org}} and the Cambridge International Dictionary of English to define the main categories involved in GLSD, as shown in Table \ref{tab2}. In addition, Table \ref{category} lists the number of images corresponding to each ship category in GLSD. The GLSD consists of a large number of ships that are capable of the oceangoing voyage (with great economic benefits), e.g., ``General cargo ship'', ``Container ship'', ``Fishing boat'', and ``Passenger ship''. Other ship types, e.g., ``Tug'', ``Canoe'', ``Sailing boat'', ``Speed boat '', and ``Barge'', are usually not capable of long-tailed travels, leading to their limited sample sizes in our dataset. Therefore, despite the involvement of additional ship categories with a significantly increased number of samples compared to other ship datasets, the class imbalance issue still exists in the proposed GLSD.
	
	\begin{table}[h]
		\centering
		\renewcommand\arraystretch{1.2}
		\caption{The number of ships corresponding to each category in GLSD.}
		\begin{tabular}{c|c}
			\hline
			Categories & Instances \\ \hline \hline
			Ore carrier            & 5,919     \\ \hline
			Bulk cargo carrier     & 16,244     \\ \hline
			General cargo ship     & 30,460     \\ \hline
			Container ship         & 29,544     \\ \hline
			Fishing boat           & 27,427     \\ \hline
			Passenger ship         & 54,206    \\ \hline
			Sailing boat           & 11,525     \\ \hline
			Barge                  & 6,099      \\ \hline
			Warship                & 13,435     \\ \hline
			Oil carrier            & 3,046      \\ \hline
			Tug                    & 3,940       \\ \hline
			Canoe                  & 3,404      \\ \hline
			Speed boat             & 7,108      \\ \hline
			Total                  & 212,357   \\ \hline
		\end{tabular}
		\label{category}
	\end{table}

	\begin{figure*}[!htb]
	\centering
	\includegraphics[height=21cm]{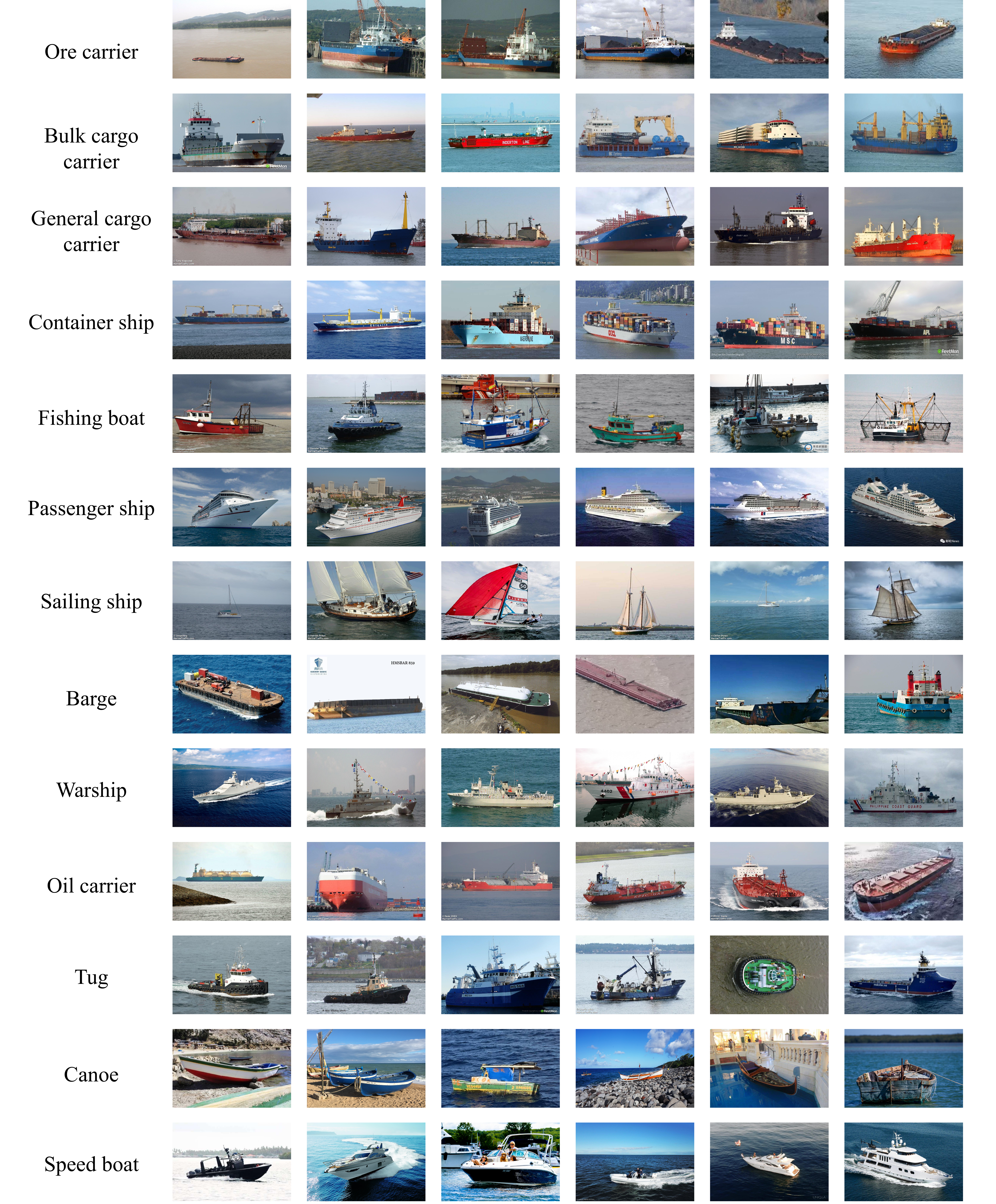}
	\caption{Samples of annotated images in the GLSD dataset. Based on the collected images, we propose 13 categories which widely exists in international routes. They are: \textsl{``Sailing boat'', ``Fishing boat'', ``Passenger ship'', ``Warship'', ``General cargo ship'', ``Container ship'', ``Bulk cargo carrier'', ``Barge'', ``Ore carrier, ``Speed boat'', ``Canoe'', ``Oil carrier'', and ``Tug''}.}
	\label{examples}
\end{figure*}
	\subsection{Statistics}
\begin{figure}[h]
		\centering
		\includegraphics[height=5.5cm]{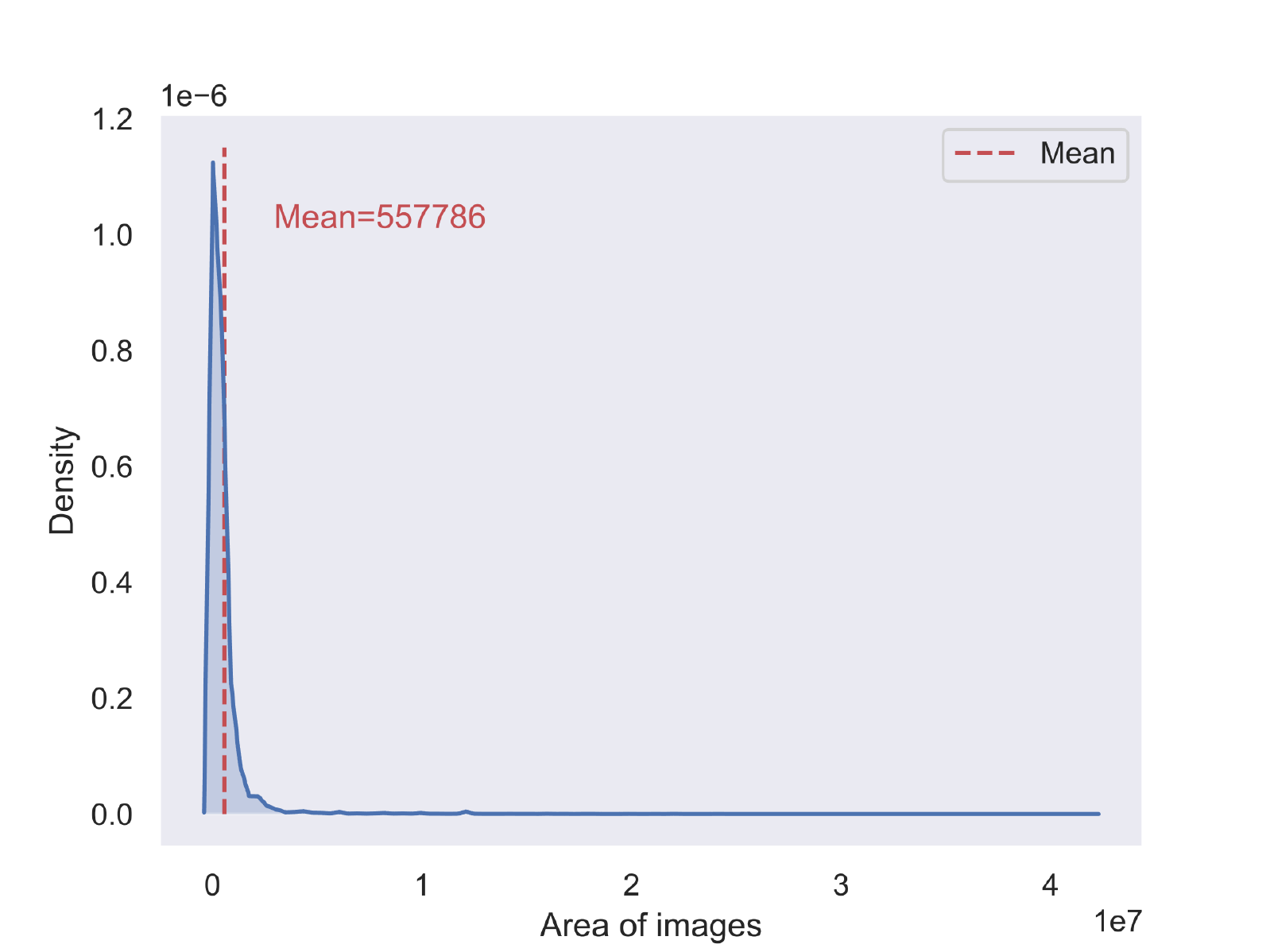}
		\caption{The distribution of the image sizes. The smallest image in this dataset has $90 \times 90$ pixels, and the largest has $6,509 \times 6,509$ pixels.}
		\label{resolution}
	\end{figure}

	Fig. \ref{resolution} shows the distribution of the image resolution in the GLSD. Different from SeaShips that mainly contain images retrieved from a monitoring system, GLSD also includes high-resolution images from unmanned aerial vehicles and satellite/airborne platforms. Considering that the performances of existing deep-learning-based algorithms are usually limited in detecting small targets, we include a large number of images with small targets (less than 32 $\times$ 32 pixels) and medium targets (between 32 $\times$ 32 to 96 $\times$ 96 pixels). The definition of small and medium targets follows \cite{lin2014microsoft}. The image sizes vary greatly in GLSD, with the smallest image of $90 \times 90$ pixels and the largest of $6,509 \times 6,509$ pixels. From the above description, it can be seen that GLSD contains more diverse images with various resolutions and target sizes than SeaShips. 
	
	\subsection{Annotation}
	In this session, we describe how we annotate images in GLSD. Different from regular objects, ships can contain other features besides their main body, such as mast, elevating equipment, and oar. During the annotation process, all object instances are labeled with object names with bounding boxes that cover the entire ship with additional ship features.

	All images are converted to ``.jpg'' format in MATLAB. We annotate images in GLSD by labelme \cite{labelme2016} following the PASCAL VOC2007 format. GLSD images in the dataset are named from ``000001.jpg'' to ``152576.jpg'', with the corresponding label files being named from ``000001.xml'' to ``152576.xml'' (we also provide the MS COCO \cite{lin2014microsoft} version of GLSD). All annotation tasks are done in a manual manner. Objects that are too dense and too small for the human eye to recognize are discarded. Fig.\ref{examples} present selected samples in the GLSD. The GLSD contains 76,189 training, 38,192 validation, and 38,195  testing images (approximately $1/2$ train, $1/4$ val, and $1/4$ test as \cite{lin2014microsoft}). We present details regarding the design of GLSD in the following sections.

	\section{Design of the GLSD dataset} \label{s4}
	
\begin{figure}[h]
		\centering
		\includegraphics[height=7.3cm]{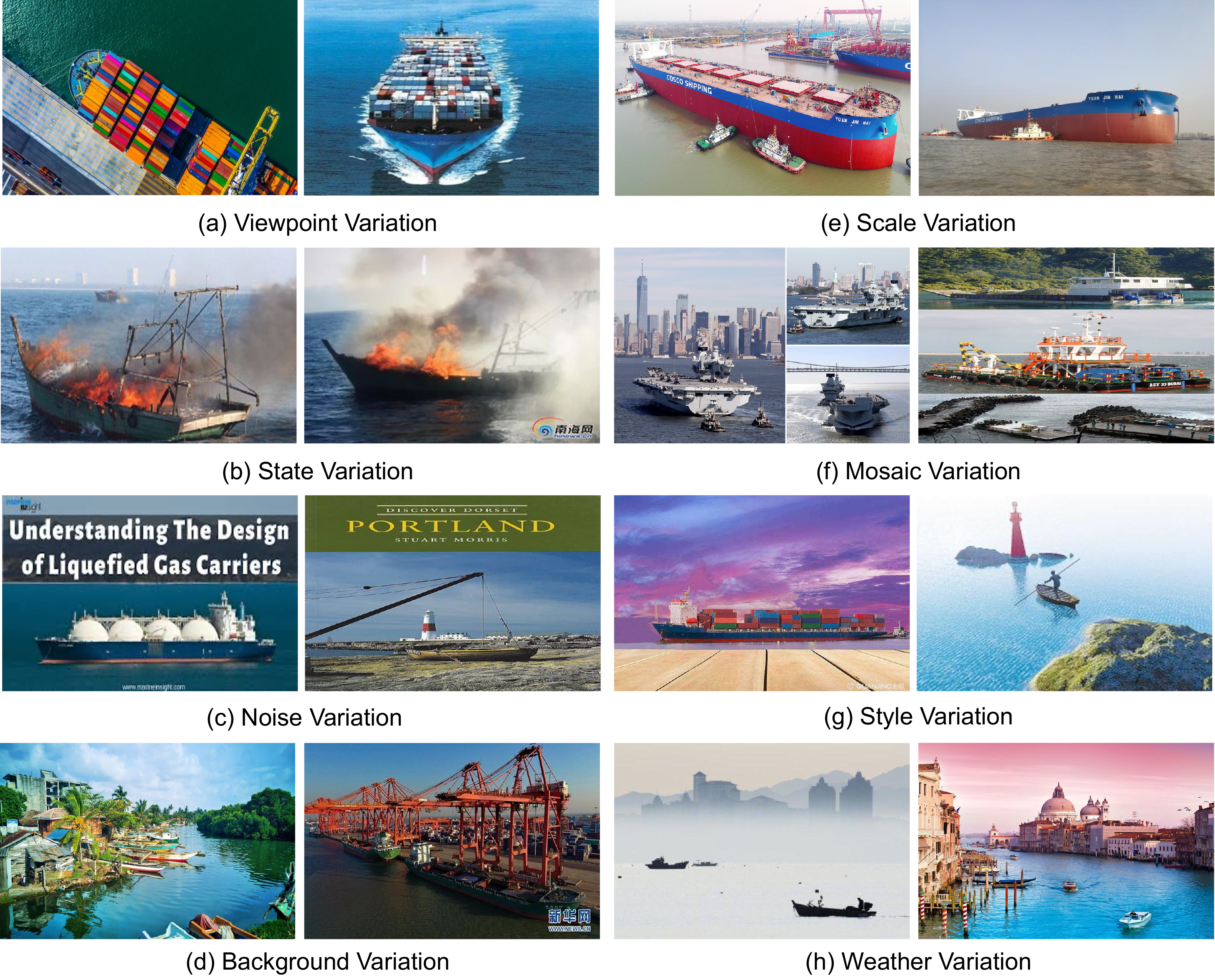}
		\caption{Selected images corresponding to different variations considered in GLSD, i.e., viewpoint, state, noise, background, scale, mosaic, style, and weather variations. }
		\label{design}
	\end{figure}

	Different from SeaShips dataset that contains images from a site monitoring system, images in GLSD collected from the Internet and searching engines are generally more complex. Eight variations, i.e., viewpoint, state, noise, background, scale, mosaic, style, and weather variations, are considered and implemented to construct the GLSD. Selected examples corresponding to these variations are outlined in Fig. \ref{design}.
	
	\subsection{Viewpoint Variation}	
	Images from different viewpoints have varying characteristics. Multi-viewpoint images have been proved to benefit deep-learning-based models in coping with the complex changes in real-world scenarios. Compared to SeaShips based on surveillance cameras with limited viewpoints, the designed GLSD contains considerably more viewpoints, as shown in Fig. \ref{design}(a), potentially leading to increased model robustness.

	\subsection{State Variation}
	SeaShips only focuses on underway ships while ignoring the state under abnormal events, such as the shipping disaster (e.g., on fire), towed by a tug, and interaction between barges and large vessels. Datasets with images under different states are beneficial in monitoring abnormal events during shipping. Fig. \ref{design}(b) shows ship images in the designed GLSD under a unique on-fire state: two fishing boats with only skeletons left after burning. 
	
		\begin{figure*}[h]
		\centering
		\includegraphics[height=15cm]{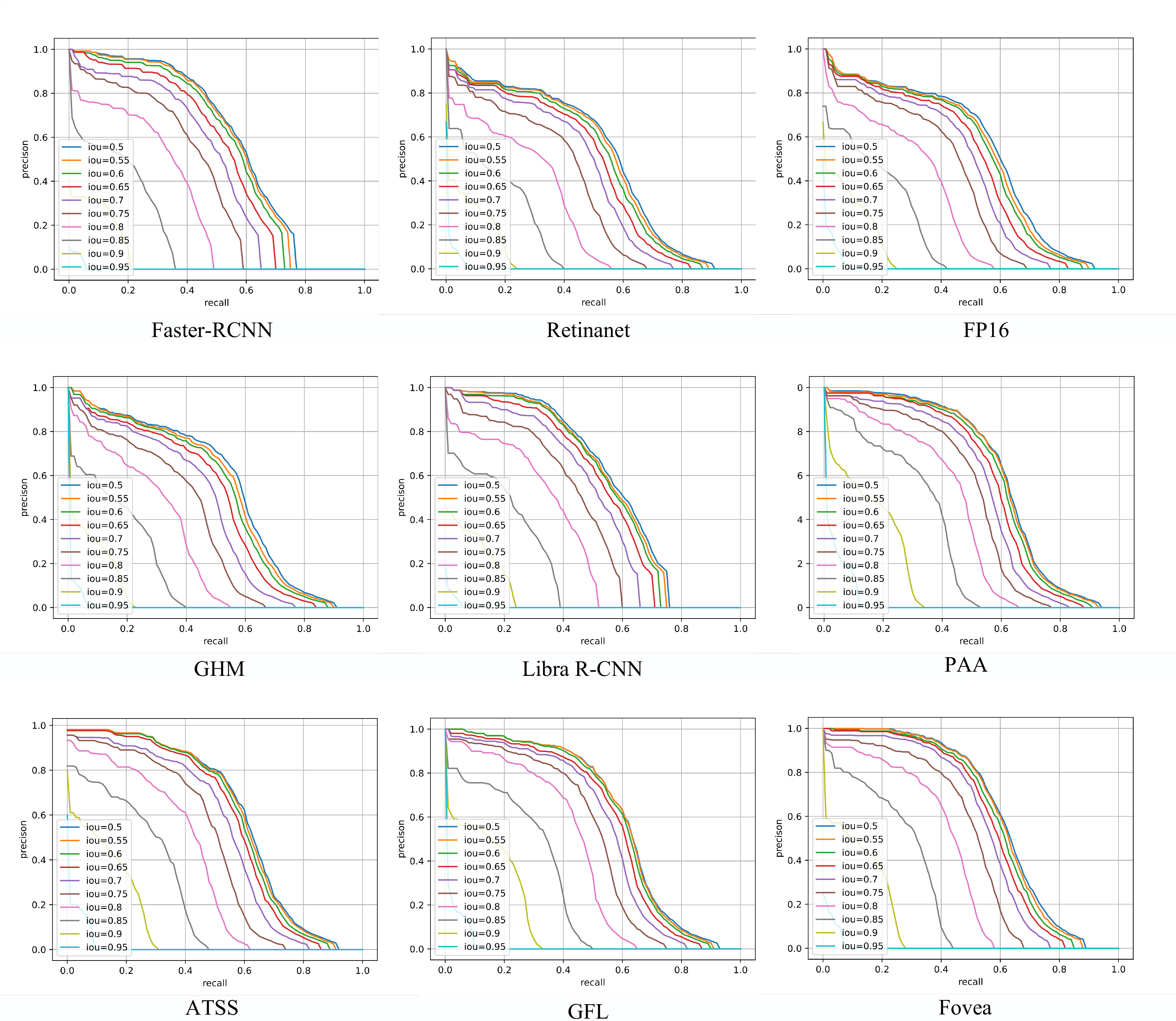}
		\caption{The Precision-Recall curve of methods with 1$\times$ schedule.}
		\label{pr}
	\end{figure*}	\subsection{Noise Variation}
	Numerous studies have shown that the accuracy of target detection is often higher in cleaner images. However, noises are unavoidable in many real-world cases. Thus, the introduction of noises in the images helps further improve the robustness of algorithms. Certain images collected from searching engines in GLSD contain watermarks (serving as noises), as shown in Fig. \ref{design}(a), (b), and (c).
	
	\subsection{Background Variation}
	Theoretically, backgrounds in a non-iconic image can provide rich contextual information. Therefore, a dataset with a diversity of backgrounds is preferred in many object recognition tasks. In real-world applications, backgrounds of the captured images tend to vary. Thus, it is necessary to collect images from different ports. Fig. \ref{design}(d) presents images with two distinctively different backgrounds: one with a tropical characteristic featured by canoes, the other one with a modern port featured by large-scale ships. 
	
	\subsection{Scale Variation}	
	Deadweight tonnage (DWT) is an indicator of the ship' sizes and transporting capacity. The DWT of ships greatly varies. For example, the maximum gross payload of an oil carrier can reach 500,000 DWT, while some old cargo ships are only with 5,000 DWT. Even for ships of the same class, the scale can considerably vary. An oil carrier can occupy ten times as many pixels as a tug, as illustrated in Fig. \ref{design}(e). Our designed GLSD contains ships with rich variations in scales, potentially increasing algorithms's capability in detecting both large and small objects. 
	
	\subsection{Mosaic Variation}
	YOLOv4 \cite{bochkovskiy2020yolov4} introduces a new method of data augmentation, named mosaicking, which mixes four different images into one image. Fig. \ref{design}(f) shows some examples of images that consist of three iconic images with different placement strategies after mosaicking. Our GLSD contains a variety of mosaicking images, greatly enriching the background information of ships to be detected.
	
		\begin{table*}[htbp]
		\centering
		\caption{Object detection results on the GLSD.}
		\renewcommand\arraystretch{1.2}
		\begin{tabular}{ccccccccccccccc}
			\toprule[1pt]
			Method & Backbone &Schedule & $AP$  & $AP_{50}$  & $AP_{75}$     & $AP_{S}$   & $AP_{M}$   & $AP_{L}$   & $AR_{1}$   & $AR_{10}$  & $AR_{100}$ & $AR_{S}$   & $AR_{M}$   & $AR_{L}$ \\
			\hline
			Faster R-CNN \cite{ren2015faster} & ResNet-50   &$1 \times$ &28.0  &44.8  &30.0   &3.4  &14.0  &31.8    &54.1    & 54.1   &54.1   &18.8   &39.2   &58.9    \\ 
			Retinanet \cite{lin2017focal} & ResNet-50  &$1 \times$&28.8  &44.4  &31.0   &3.6  &14.5  &32.7    &62.6    &62.6    &62.6   &28.9   & 49.1  & 67.9   \\ 
			FP16 \cite{micikevicius2017mixed} & ResNet-50  &$1 \times$ & 29.2 &44.9  &31.5   &3.3  &14.7  &33.3    &62.9    &62.9    &62.9   &29.4   &49.4   &68.2    \\ 
			GHM \cite{li2019gradient} & ResNet-50 &$1 \times$  &28.6  &44.3  &30.6   &3.4  &14.4  &32.6    &62.2    &62.2    &62.2   & 27.2  &48.5   &67.7   \\
			Libra R-CNN \cite{pang2019libra} & ResNet-50 &$1 \times$  &28.2  &44.6  &30.6   &3.4  &13.9  & 32.3   &54.4    &54.4    &54.4   &21.6   &39.9   &59.0  \\
			PAA \cite{kim2020probabilistic}& ResNet-50 &$1 \times$  &30.9  &45.8  &33.9   &3.9  &15.6  &35.2    &67.2    &67.2    &67.2   &34.3   &53.7   &72.5    \\  
			ATSS \cite{zhang2020bridging}& ResNet-50 &$1 \times$  &30.1  &45.3  &32.7   &3.6  &15.4  &34.2    &65.3   &65.3    &65.3   &32.1   &51.5   &70.7    \\   
			GFL \cite{li2020generalized} & ResNet-50 &$1 \times$   &30.9  & 45.9 & 33.7  &3.5  &15.6  &35.1    & 65.5   &65.5    & 65.5  &30.3   &51.4   &71.7    \\ 
			Fovea \cite{kong2020foveabox} & ResNet-50 &$1 \times$  &30.1  &46.5  &32.4   &3.7  &15.4  & 34.2   &48.6    & 60.4   &60.9   &27.1   &46.3   &66.2    \\ \hline
			Faster R-CNN \cite{ren2015faster} & ResNet-50   &$2 \times$&29.3  & 46.1 &   32.0 &3.7  &14.9  & 33.5  &54.5    & 54.5   &54.5   &19.1   &38.7   & 59.7   \\ 

			Retinanet \cite{lin2017focal} & ResNet-50 &$2 \times$  &29.5  &45.3  &31.8   &3.5  &14.6  &33.8    &59.5    &59.5    &59.5   &23.1   & 44.4  & 65.3   \\ 
			PAA \cite{kim2020probabilistic}& ResNet-50 &$2 \times$  &31.9  &47.1  &34.9   &3.7  & 16.1 &36.3    &50.2    &65.2    &66.0   &32.0   &51.7   & 71.6    \\ 
			\toprule[1pt] 
		\end{tabular}%
		\label{result1}%
	\end{table*}%

		\begin{table*}[htbp]
		\centering
		\caption{Detection performance evaluated on the GLSD.}
		\renewcommand\arraystretch{1.2}
		\begin{tabular}{cccccccccc}
			\toprule[1pt]
			Method & Faster R-CNN \cite{ren2015faster} & Retinanet \cite{lin2017focal}   & FP16 \cite{micikevicius2017mixed}  & GHM \cite{li2019gradient}     & Libra R-CNN \cite{pang2019libra}   & PAA \cite{kim2020probabilistic}  & ATSS \cite{zhang2020bridging}   & GFL \cite{li2020generalized}   & Fovea \cite{kong2020foveabox}  \\
			\hline
			Ore carrier  &35.8  &32.2 &33.9  &33.4  &36.8   &44.1   &40.9   &42.6    &42.2  \\ 
			Bulk cargo carrier &17.7  &16.4 &16.2  &16.3  &18.2   &21.0   &18.5   &19.9    &18.8   \\ 
			General cargo ship &24.6  &24.9 &25.2  &25.6  &24.8   &28.0   &26.7   &27.7    &26.4   \\ 
			Container ship &57.8  &62.0 &62.4  &62.2  &58.6   &63.1   &63.3   &64.1    &62.9   \\
			Fishing boat &24.1  &26.6 &26.7  &26.7  &24.2   &28.0   &26.6   & 27.7   &26.6   \\
			Passenger ship&50.3  &52.1 &52.2  &50.9  &50.7   &53.4   &52.8   &53.2    &51.5    \\  
			Sailing boat&47.0  &48.3 &48.5  &48.2  &47.0   &49.5   &48.5   &48.9    &47.8   \\   
			Barge &5.0  &3.8 &4.4  &4.2  & 5.3  &5.4   & 4.8  &5.7    &5.2       \\ 
			Warship &42.7  &44.6 &45.2  &44.6  &43.2   & 46.7  & 45.7  &46.7    & 45.0   \\
			Oil carrier &17.4  &20.0 &20.5  &16.5  & 17.0  &  20.0  &18.1   &21.2    & 19.9   \\ 
			Tug &19.7  &19.0 &19.8  &19.5  &19.3   &  21.4  &20.5   &19.6    & 20.5   \\ 
			Canoe &12.3  &13.3 &13.3  &13.7  &12.1   &12.8  &14.2   &13.7    &13.6     \\ 
			Speed boat &9.2  &10.8 &11.3  &10.3  &9.2   &11.3   &10.5   &11.0    &10.5    \\ \hline
			\toprule[1pt] 
		\end{tabular}%
		\label{cla_ap}%
	\end{table*}
	In addition to multi-viewpoint images, our designed GLSD includes images from various categories: aerial images, remote sensing images, and portraits. Numerous efforts have been made towards style transfer as a data augmentation approach (e.g., domain adaptation between GTA5 and image style transfer on the COCO database). Our GLSD contains abundant image styles that include images captured via cameras and realistic paintings, as shown in Fig. \ref{design}(g).

	\subsection{Weather Variation}
	It is widely acknowledged that port operations are susceptible to extreme weather conditions, such as high winds, fog, heavy haze, snowstorms, thunderstorms, and typhoons. Such extreme weather conditions greatly affect the arrival and departure of ships and the unloading of cargo in the port. On the sea, the weather tends to change significantly in a relatively short time. Our DLSD includes a variety of weather conditions, expected to benefit models in ship recognition under different weather scenarios, as illustrated in Fig. \ref{design}(h).

	In summary, the aforementioned variations make GLSD a rather challenging dataset for ship detection and recognition. The rich variations as well as the effectively widening within-class gap in our GLSD are expected to facilitate models in reaching higher robustness.


	\section{Evaluation Results}\label{s6}

	\subsection{Baseline Algorithms}
	In this section, we conduct a comprehensive comparison of the following state-of-the-art object detection algorithms on GLSD: Faster R-CNN \cite{girshick2015region}, RetinaNet \cite{lin2017focal}, GHM \cite{li2019gradient}, FP16 \cite{micikevicius2017mixed}, Libra R-CNN \cite{pang2019libra}, PAA \cite{kim2020probabilistic}, ATSS \cite{zhang2020bridging}, GFL \cite{li2020generalized}, and Fovea \cite{kong2020foveabox}.

	\subsection{Implementation Details}
	
	 These experiments run at a desktop based on mmdetection-2.12.0 \cite{mmdetection} (a popular open-source object detection toolbox developed by OpenMMLab)\footnote{\url{https://github.com/open-mmlab/mmdetection}} with three NVIDIA GTX TITAN GPUs and 3.60 GHz Intel Core i7-7820X CPU, 32GB memory. We implement these methods using the PyTorch 1.7.0 \cite{paszke2019pytorch} library with Python 3.7.9 under Ubuntu 18.04, CUDA 10.2, and CUDNN 7.6 systems.

	For evaluation, we employed average precision ($AP$, $AP_{50}$, $AP_{75}$, $AP_{S}$, $AP_{M}$, and $AP_{L}$) and average recall ($AR_{1}$, $AR_{10}$, $AR_{100}$, $AR_{S}$, $AR_{M}$, and $AR_{L}$), as with \cite{zhang2020object}. Among them, the $AP$, $AP_{S}$, $AP_{M}$, $AP_{L}$, and all average recall are calculated with intersections over union (IOU) values ([0.50 : 0.05 : 0.95]) as IOU thresholds. As for $AP_{50}$ and $AP_{75}$, the corresponding thresholds are 0.5 and 0.75, respectively. Moreover, $ scale = {\left \{ S, M, L\right \} }$ represents the average with different scales (small scale: targets with less than 32 $\times$ 32 pixels; medium scale: targets with between 32 $\times$ 32 to 96 $\times$ 96; large scale: targets with larger than 96 $\times$ 96 pixels \cite{lin2014microsoft}), and $num = {\left \{ 1, 10, 100\right \} }$ denotes the average recall with different number of detections.

	\subsection{Results and Analysis}
	For a fair comparison, all selected algorithms are trained and tested on images with 1,333 $\times$ 800 pixels. In Table \ref{result1}, we report the performances of all selected models on GLSD. The prediction-recall curves are shown in Fig.\ref{pr}. In scenes that contain small targets, APs from selected algorithms without the focal loss function are lower than 5\%. Even in scenes with medium targets, APs of all selected algorithms with schedule 1$\times$ are about 16\%, proving that small-target recognition is still one of the major challenges in our designed GLSD. Thus, we believe our GLSD creates a valuable venue for future innovative object detectors to compete.

	With the introduction of the focal loss, an effective approach to mitigating the issues of long-tailed distribution, the performances of Retinanet \cite{lin2017focal} and GFL \cite{li2020generalized} show significant improvement compared to other two-stage algorithms. We notice that ARs are larger than APs for these algorithms, indicating the existence of error detection due to the small inter-class gap. However, with an increasing number of iterations, Retinanet \cite{lin2017focal} is able to achieve a great performance (up to 1.0\% gains) in large-object detection. Our experiment suggests that solutions that address the long-tailed distribution are the key for models to reach satisfactory performance in GLSD.
	
	The prediction-recall curves of PAA \cite{kim2020probabilistic} with different training schedules are shown in Fig.5. We observe notable increases in AP (about 1.0\% gains) and stableness in AR. With the increase of IOU, the impact of the number of iteration on performance becomes notable, especially when $iou \ge 0.8$. It implies that with the increase of iterations leads to improved capability of the model in identifying non-ship objects. 
	
	In order to further investigate the performance of different classes on the GLSD, Table \ref{cla_ap} shows the APs of the above state-of-the-art object detection algorithms. As shown in Table \ref{cla_ap}, the APs of ``Barge'' are only about 5\%, presumably due to two reasons: 1) the small number of ``Barge'' images; 2) the similarity between ``Barge'' and ``Ore carrier'' (especially in shape). The same phenomenon is observed in other categories with a small number of images (``Oil carrier'', ``Tug'', ``Canoe'', and ``Speed boat''). However, the tested methods all have a good recognition performance in ``Warship'',  given their unique appearances.

	\begin{figure}[h]
		\centering
		\includegraphics[height=4cm]{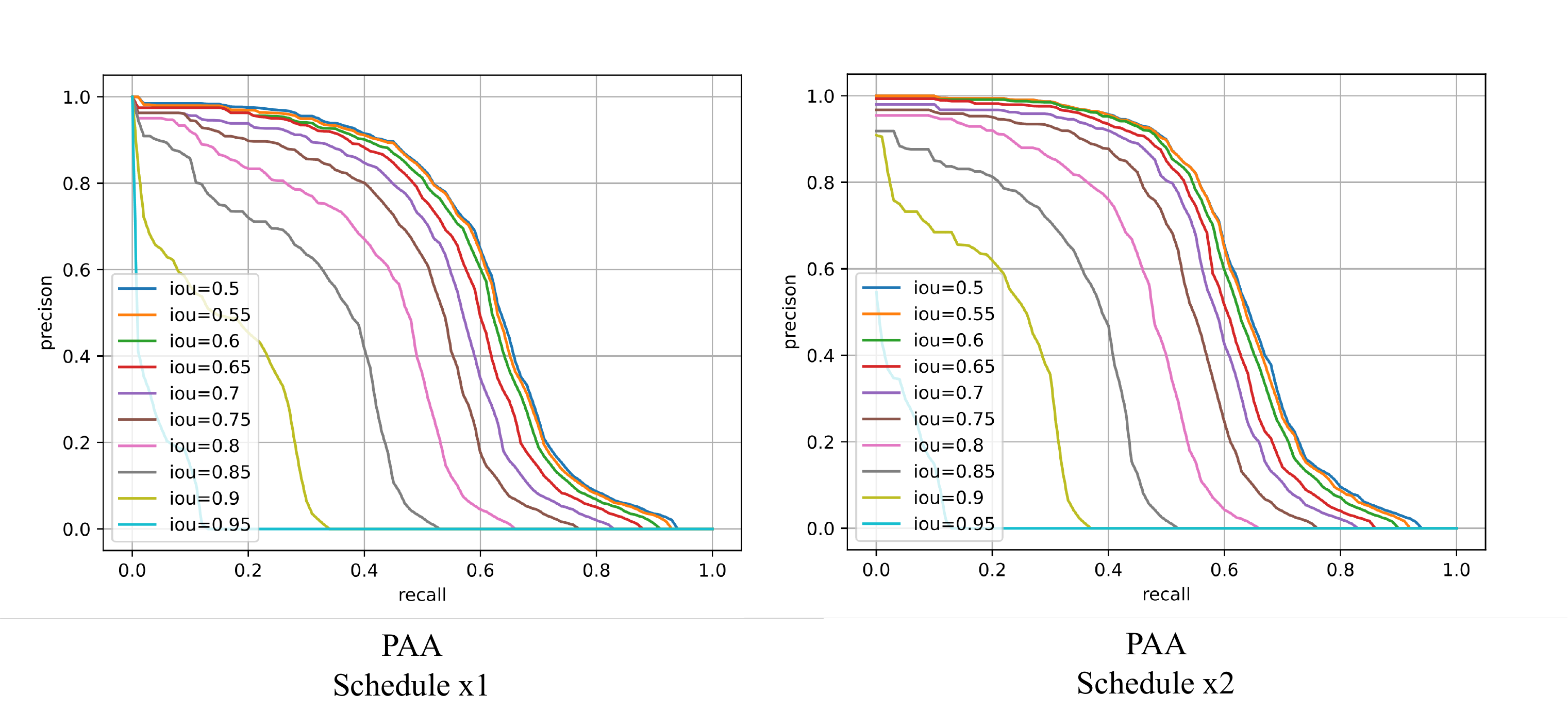}
		\caption{The Precision-Recall curves of PAA with different training schedule.}
		\label{sch}
	\end{figure}
	
	\subsection{Ablation Studies}
	
	1) Validation of multi-scale training: In our detection framework, all images are resized to 1,333 $\times$ 800 pixels for training. Considering the poor performance of small target detection on GLSD, we train the state-of-the-art method (PAA \cite{kim2020probabilistic}) with default multi-scale configuration (1,333 $\times$ 800 and 1,333 $\times$ 640 pixels) to verify the impact of multi-scale training on GLSD. As shown in Table \ref{ms}, different configurations of the training size lead to trivial performance fluctuations. Such a phenomenon can be explained by the fact that there exist huge differences in terms of image sizes in GLSD (some natural images with 96 $\times$ 96 pixels and satellite images with 6509 $\times$ 6509 pixels), while most images in regular natural image datasets (e.g., PASCAL VOC2007 \cite{everingham2010pascal} and COCO \cite{lin2014microsoft}) are no more than 1000 $\times$ 1000 pixels.

	\begin{table}[h]
		\centering
		\caption{Ablation studies of training image sizes.}
		\renewcommand\arraystretch{1.2}
		\begin{threeparttable}
	\begin{tabular}{cccccc}
	\toprule[1pt]
\multirow{2}{*}{Method} & \multirow{2}{*}{PAA-s1} & \multirow{2}{*}{PAA-s2} & \multicolumn{3}{c}{MS PAA} \\
 &  &  & s1 & s2 & MS \\ \hline
	$AP$ &30.9  &31.1   &30.5   &30.5  &30.5   \\ \hline
	$AP_{50}$ &45.8  &46.1  &45.1 &45.1  &45.1  \\ \hline
	$AP_{75}$ &33.9  &34.2  &33.5  &33.6  &33.6 \\ \hline
	$AP_{S}$ &3.9  &3.7  &3.7  &3.9  &3.9 \\ \hline
	$AP_{M}$ &15.6  &15.5  &14.9  &14.9  &15.6 \\ \hline
	$AP_{L}$ &35.2  &35.4  &34.6 &34.7  &34.7  \\ \hline
	$AR_{1}$ &67.2  &65.7  &67.2  &67.2  &67.6 \\ \hline
	$AR_{10}$ &67.2  &66.1  &67.2 &67.2  &67.6  \\ \hline
	$AR_{100}$ &67.2  &66.9  &67.2 &67.2  &67.6  \\ \hline
	$AR_{S}$ &34.3  &33.2  &33.3  &33.3  &35.8 \\ \hline
	$AR_{M}$ &53.7  &52.9  &53.9 &53.9  &55.2  \\ \hline
	$AR_{L}$ &72.5  &72.4  &72.7 &72.7  &72.7  \\ \toprule[1pt]
	\end{tabular}
\label{ms}
		\begin{tablenotes}
		\footnotesize
		\item s2: 1,333 $\times$ 800 images
		\item s1: 1,333 $\times$ 640 images
		\item MS PAA : 1,333 $\times$ 640 and 1,333 $\times$ 800 images
	\end{tablenotes}
\end{threeparttable}
	\end{table}

	2) Validation of normalization strategies: The normalization strategy in the detection task can effectively accelerate the convergence speed and alleviate the problem of gradient disappearance. To intuitively reveal the impact of normalization on GLSD, different normalization methods (Batch Normalization (BN) \cite{ioffe2015batch}, Group Normalization (GN) \cite{wu2018group}, and Synchronized Batch Normalization (SyncBN) \cite{zhang2018context}) on PAA \cite{kim2020probabilistic} are tested. The GN layer is proposed to eliminate the influence of batch size for normalization, while the SyncBN layer is distributed version BN layer. As shown in Table \ref{norm}, the AP of PAA with SyncBN is 0.3\% and 3.6\% higher than that with BN and the GN, respectively.

	\begin{table}[h]
		\centering
		\caption{Ablation studies of normalization strategies.}
		\renewcommand\arraystretch{1.2}
			\begin{threeparttable}
	\begin{tabular}{c|ccc}
	\toprule[1pt]
	Method & PAA w/ BN & PAA w/ GN & PAA w/ SyncBN \\ \hline
	$AP$ &30.9  &27.6  & 31.2 \\ \hline
	$AP_{50}$ &45.8  &41.4  &46.3  \\ \hline
	$AP_{75}$ &33.9  &30.1  & 34.2 \\ \hline
	$AP_{S}$ &3.9  &3.2  &3.5  \\ \hline
	$AP_{M}$ &15.6  &13.1  &15.9  \\ \hline
	$AP_{L}$ &35.2  &31.4  &35.5  \\ \hline
	$AR_{1}$ &67.2  &66.0  & 67.0 \\ \hline
	$AR_{10}$ &67.2  &66.0  &67.0  \\ \hline
	$AR_{100}$ &67.2  &66.0  &67.0  \\ \hline
	$AR_{S}$ &34.3  &31.6  &34.0  \\ \hline
	$AR_{M}$ &53.7  &52.0  &53.6  \\ \hline
	$AR_{L}$ &72.5  &71.6  &72.3  \\ \toprule[1pt]
	\end{tabular}
	\label{norm}
	\begin{tablenotes}
		\footnotesize
		\item w/: with
		\item BN: Batch Normalization
		\item GN: Group Normalization
		\item SyncBN: Synchronized Batch Normalization
	\end{tablenotes}
\end{threeparttable}
	\end{table}

	\section{Conclusion} \label{s7}
	
	In this paper, we introduce a global large-scale ship database, i.e., GLSD, which is designed for ship detection tasks. The designed GLSD is considerably larger and more challenging than any existing database, to our best knowledge. The main characteristics of the GLSD lie in three aspects: 1) the GLSD contains a total of 152,576 images with a widening inter-class gap from 13 categories, i.e., ``sailing boat'', ``fishing boat'', ``Passenger ship'', ``Warship'', ``General cargo ship'', ``Container ship'', ``Bulk cargo carrier'', ``Barge'', ``Ore carrier, ``Speed boat'', ``Canoe'', ``Oil carrier'', and ``Tug''; 2) the GLSD includes a diversity of variations that include viewpoint, state, noise, background, scale, mosaic, style, and weather variations, which benefit improved model robustness; 3) the route-based version of GLSD, i.e., GLSD\_port, contains geographic information, providing rich multi-modal information that benefits various ship detection and recognition tasks. We also propose evaluation protocols and provide evaluation results on GLSD using numerous state-of-the-art object detection algorithms. As ship images of certain categories are difficult to collect, the current version of GLSD has a notable long-tail issue. We will continue to extend GLSD with more ship images, especially on ship categories of ``Tug'', ``Canoe'', and ``Speed boat''.

	\section*{Acknowledgement}
	
	We thank Lan Ye, Sihang Zhang, Linze Bai, Gui Cheng, and all the others who were involved in the annotation of GLSD. In addition, we thank the support of the Post-Doctoral Research Center of Zhuhai Da Hengqin Science and Technology Development Co., Ltd, Guangdong Hengqin New Area.
	\bibliographystyle{IEEEtran}
	\bibliography{IEEEabrv,dataset}

\end{document}